\documentclass[10pt,twocolumn,letterpaper]{article}

\usepackage{iccv}
\usepackage{times}
\usepackage{epsfig}
\usepackage{graphicx}
\usepackage{amsmath}
\usepackage{amssymb}

\usepackage{color}
\usepackage{booktabs}
\usepackage{lipsum}
\usepackage{color}
\usepackage[normalem]{ulem}

\usepackage{multirow}
\usepackage[table,xcdraw]{xcolor}

\usepackage[breaklinks=true,bookmarks=false]{hyperref}

\iccvfinalcopy 

\def\iccvPaperID{****} 

\ificcvfinal\fi

\begin{document}

\title{\ A Multi-modal and Multi-task Learning Method for Action Unit and Expression Recognition}

\author{Yue Jin\footnotemark[1]\\
China Pacific Insurance (Group) Co., Ltd.\\
719 Yishan Road, Shanghai, China\\
{\tt\small jinyue-007@cpic.com.cn}
\and
Tianqing Zheng\footnotemark[1]\\
Shanghai Jiaotong University\\
800 Dongchuan Road, Shanghai, China\\
{\tt\small sjtuztq@sjtu.edu.cn}
\and
Chao Gao\\
China Pacific Insurance (Group) Co., Ltd.\\
719 Yishan Road, Shanghai, China\\
{\tt\small gaochao-027@cpic.com.cn}
\and
Guoqiang Xu\\
China Pacific Insurance (Group) Co., Ltd.\\
719 Yishan Road, Shanghai, China\\
{\tt\small xuguoqiang-009@cpic.com.cn}
}

\maketitle
\ificcvfinal\thispagestyle{empty}\fi

\renewcommand{\thefootnote}{\fnsymbol{footnote}}
\footnotetext[1]{These authors contributed equally to this work.}

\begin{abstract}
Analyzing human affect is vital for human-computer interaction systems.
Most methods are developed in restricted scenarios which are not practical
for in-the-wild settings. The Affective Behavior Analysis in-the-wild (ABAW)
2021 Contest provides a benchmark for this in-the-wild problem.
In this paper, we introduce a multi-modal and multi-task learning method
by using both visual and audio information.
We use both AU and expression annotations to train the model and apply
a sequence model to further extract associations between video frames.

We achieve an AU score of 0.712 and an expression score of 0.477 on the validation set. 
These results demonstrate the effectiveness of our approach in improving model performance.
\end{abstract}

\section{Introduction}

Nowadays, analyzing human affect is becoming more and more important for
Artificial Intelligence (AI) systems,
especially for human-computer interaction.
The ability for machines to recognize the
human face is mature, but the ability to understand human emotions still has a long way to go.

The Affective Behavior Analysis in-the-wild (ABAW2) 2021 Competition \cite{2106.15318}\cite{kollias2019deep}\cite{kollias2019expression}
\cite{kollias2019face}\cite{kollias2020analysing}\cite{kollias2021affect}
\cite{kollias2021distribution}\cite{zafeiriou2017aff} is held by
Kollias et al. in conjunction with ICCV 2021.
It provides a benchmark for three main tasks of Valence-
Arousal Estimation, seven Basic Expression Classification
and twelve Action Unit Detection\cite{ekman1997face}.
The Facial Action Coding System (FACS) is a comprehensive system for describing
facial movement. Action Units (AU) are individual components of muscle movements\cite{ekman1997face}.

\begin{figure}[t]
\begin{center}
\includegraphics[width=\linewidth]{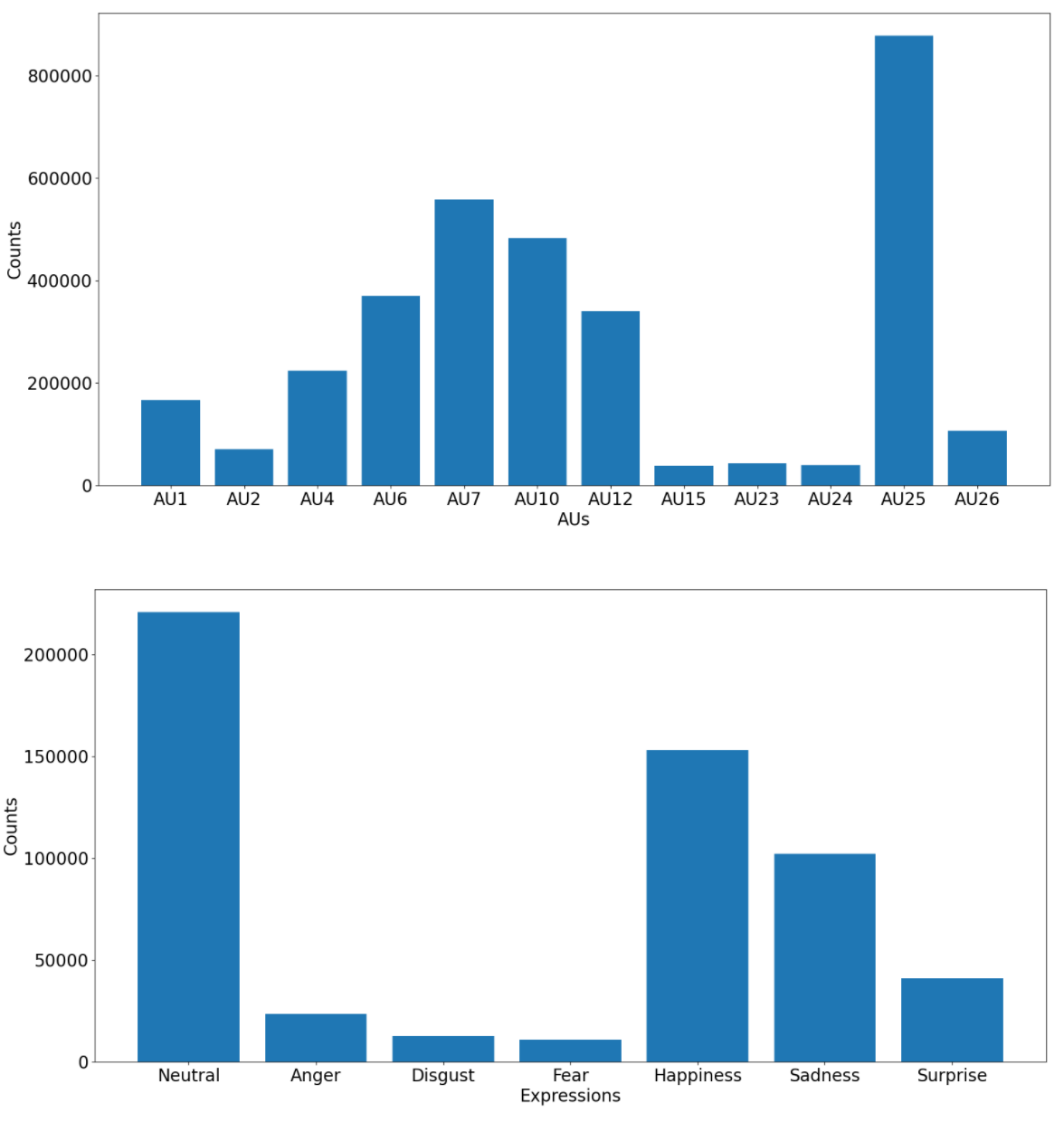}
\end{center}
   \caption{Statistics on the number of AUs and expressions labeled in each category.}
\label{fig:stat au expr}
\end{figure}

All three tasks are based on a large-scale in-the-wild database,
Aff-Wild2\cite{kollias2019expression}\cite{kollias2018aff}.
It consists of 548 videos with 2,813,201 frames and provides annotations
for all of these tasks. All videos are collected from Youtube which provides
a real-world setting. So it is much more difficult to analyze
affect than for other datasets. Data imbalance issues make this competition
quite challenging. See figure.\ref{fig:stat au expr}.

To tackle these problems, we propose a multi-task method to learn
Action Unit (AU) and facial expressions jointly
using both visual and audio information. First, we train a visual model.
Second, we freeze the parameters of the visual model and train the audio model.
Third, we concatenate both visual and audio features and train the sequence model.

\begin{figure*}[t]
\begin{center}
\includegraphics[width=\linewidth]{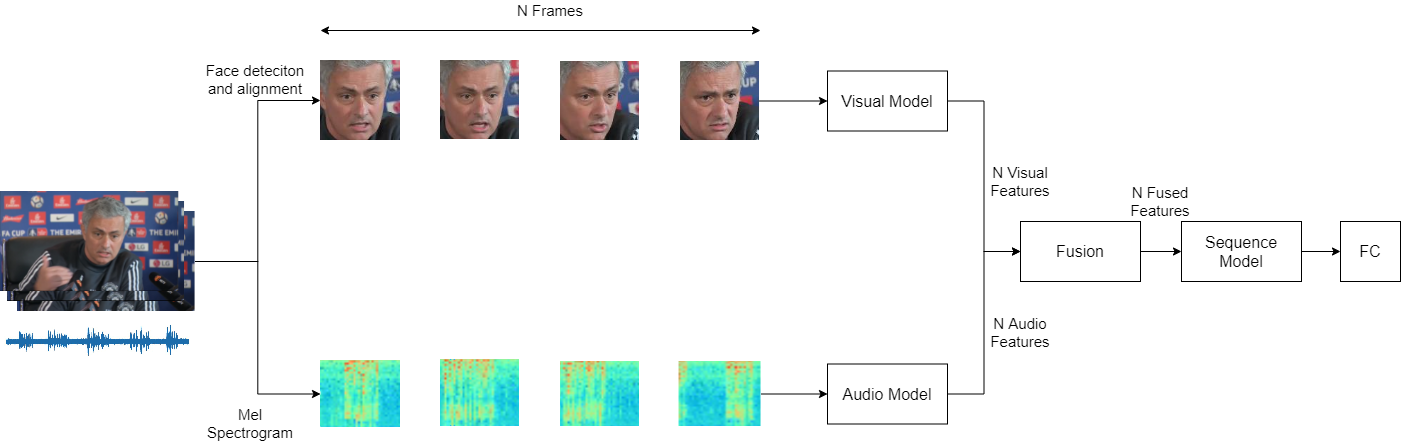}
\end{center}
   \caption{A multi-modal framework.}
\label{fig:multimodal}
\end{figure*}

\section{Related Work}
In the first ABAW2 competition, lots of
teams presented great methods for this challenging problem. 

Deng et al.\cite{deng2020multitask} proposed a multi-task learning method
to learn from missing labels. They used a data balancing technique to
the dataset. First, they used the ground truth labels of all three tasks
to train a teacher model. Secondly, they used the output of the teacher
model as the soft labels. They used the soft labels and the ground truth
labels to train their student models.

Kuhnke and Rumberg\cite{kuhnke2020two} proposed a two-stream aural-visual
model. Audio and image streams are first proposed separately and fed into a CNN
network. Then they use temporal convolutions to the image stream. They use
additional features extracted during facial alignment and correlations
between different emotion representations to boost their performance.

\section{Method}

We use a multi-modal multi-task learning method for facial action units and expression recognition. See Figure.\ref{fig:multimodal}

First, we train the visual model separately
by using both AU and expression annotations. Secondly, we freeze the
parameters of the visual stream and add an audio stream to extract the audio feature.
Finally, the visual feature and the audio feature are concatenated and fed into
a transformer encoder to further extract temporal features.

\begin{figure*}[t]
\begin{center}
\includegraphics[scale=0.55]{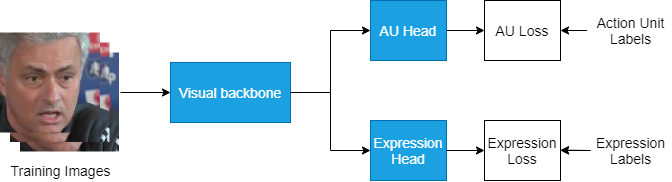}
\end{center}
   \caption{A multi-task visual stream training framework.}
\label{fig:visual}
\end{figure*}

\subsection{Multi-Task Visual model}

A multi-task framework is used to train the visual model, see Figure.\ref{fig:visual}.

First, we train the visual backbone with Cosface loss.\cite{wang2018cosface}
The dataset we use is Glint360K.\cite{an2020partial} It is the largest
and cleanest face recognition dataset, which contains 17,091,657 images of
360,232 individuals. Using the pretrained visual backbone boosts
the performance because it provides sufficient human face information.

Both AU and expression heads share weights of the same backbone. 
AU and expression heads are fully connected layers that map
features to the number of output classes.
The annotations of AU and expression is incomplete, that is,
some frames have both AU and expression annotation, but other
frames only have one of these two annotations.
To tackle this problem, we design two ways to do backpropagation when training.

1) Epoch by epoch:
The parameters of the AU head and expression head are updated in rotation
epoch by epoch. For example, we have a set of images, annotations
of AU, and annotations of expression.
At epoch 1, we use images with expression annotations, do backpropagation,
and only update the parameters of backbone and expression head.
At epoch 2, we use images with AU annotations, do backpropagation,
and only update the parameters of the backbone and AU head.
Then we repeat the above steps.

2) Batch by batch:
The parameters of the AU head and expression head are updated in rotation
batch by batch. For example,
for the first batch, we use images with expression annotations, do backpropagation,
and only update the parameters of the backbone and expression head.
For the next batch, we use images with AU annotations, do backpropagation,
and only update the parameters of the backbone and AU head.
Then we repeat the above steps.

\subsection*{Loss Function}

For action unit recognition, we use BCE loss with position weight to tackle the imbalance of positive samples and negative samples.
It's possible to trade-off recall and precision by adding weights to positive examples.



\begin{equation}
L_{BCE} = \mathbb{E}[-\sum (w_{i}t_{i} \cdot log p_{i} + (1-t_{i})\cdot log(1-p_{i}))] \label{XX}
\end{equation}

For expression recognition, we use focal loss\cite{lin2017focal} to
tackle class imbalance problem.
It uses a modulating factor to the cross-entropy loss to reduce
the loss contribution from easy examples
and elevate the importance of hard examples.

\begin{equation}
FL(p_{t}) = - \alpha_{t}(1-p_{t})^{\gamma}log(p_{t}) \label{XX}
\end{equation}



\subsection{Multi-Modal Sequence Model}
The Multi-modal Sequence model is composed of two sub-modal modules(Visual Model and Audio Model), a fusion layer, and an encoder layer from transformer\cite{vaswani2017attention}.

The input of the network is a video and the output is the predicted labels of every frame in the video.

First, image frames and the audio stream
are extracted from the video and fed into the visual and audio models respectively. 

\textbf{Visual model:} 
The Visual Model is pretrained using the multi-task method in Section 3.1 to extract visual features of a single frame.

\textbf{Audio model:} 
For audio, a Mel spectrogram is computed using the TorchAudio package and  TDNN\cite{peddinti2015time} is used as the backbone to extract the audio features.

\textbf{Sequence model:} 
Then, the audio features and the visual features are aligned and fused to get the multi-model features. The multi-model features are fed into the encoder network from transformer\cite{vaswani2017attention} to extract the sequence features; 

Finally, a fully connected layer is used to get the prediction result of each frame.




\section{Experiments}

\subsection{Dataset}

Aff-wild2 dataset\cite{kollias2019expression}\cite{kollias2018aff} is used
for both AU and expression recognition. We discard annotations with -1.

For AU recognition, we use BP4D\cite{zhang2014bp4d} as an additional training dataset.
This replenishes the number of scarce categories, like AU15, AU23, AU24.

Because some of the images in the AU and expression validation sets appear in each other's training sets, we remove this part of the images that appear in the expression training set when validating the AU metric to ensure that there is no additional prior knowledge when validating the AU metric. The corresponding operation is also performed when verifying the expression metrics.
Thus, we ensure that the validation set videos in AU and expression recognition
tasks are consistent and the validation set videos of each task do not appear
in the training set of the other task.

We use the cropped and aligned images provided in the Aff-wild2 dataset.

\subsection{Experiment Settings}

Our framework is implemented by using Pytorch\cite{paszke2019pytorch}. 

\textbf{Visual model setting:}
The face recognition model - IResNet100\cite{he2016deep} provided by Insightface\footnote{https://github.com/deepinsight/insightface}
is used as the pretrained model. Other visual backbones, such as
SENet\cite{hu2018squeeze} et.al. are also trained with Cosface loss\cite{wang2018cosface} using Glint360K\cite{an2020partial} dataset.

We use the cropped and aligned images provided by Aff-wild2 dataset.
The width and height are set to 112 pixels.
Data augmentations are random horizontal
flip, small random crop, and small random changes to hue, saturation, and lightness. The mini-batch size is set to 64.
We use SGD\cite{ruder2016overview} optimizer with momentum and the learning
rate is set to 0.001.

\textbf{Audio model setting:}
We use the following settings to compute a Mel spectrogram of the audio:
\begin{itemize}
\item{number of mel filter banks $n_{mels} = 64$}
\item{window size $w_{win} = 10 ms$}
\item{window stride $t_{stride} = 5 ms$}
\end{itemize}

The output dimension of Time Delay Neural Network (TDNN\cite{peddinti2015time})is set to 512.

\textbf{Sequence model setting:}
The input number of frames is set to 30 and the number of encoder layers from the transformer is set to 1.
We use SGD\cite{ruder2016overview} optimizer with momentum and the learning
rate is set to 0.01.

\subsection{Evaluation Metric}

For 12 Action Unit Detection, the performance metric is:
\begin{equation}
0.5 * F1 \underline{~~} Score + 0.5 * Accuracy
\end{equation}

For 7 Basic Expression Classification, the performance metric is: 
\begin{equation}
0.67 * F1 \underline{~~} Score + 0.33* Accuracy
\end{equation}
where F1 Score is the unweighted mean and Accuracy is the total accuracy.


\begin{table}[t]
\caption{AU recognition results on validation dataset}
\begin{center}
\begin{tabular}{|l|c|c|c|}
\hline
Method & Average F1 & Total Accuracy & Metric \\
\hline
Baseline\cite{2106.15318} & 0.40 & 0.22 & 0.31 \\
Ours & 0.545 & 0.879 & 0.712\\
\hline
\end{tabular}
\end{center}
\label{tab:au}
\end{table}

\begin{table}[t]
\caption{Expression recognition results on validation dataset}
\begin{center}
\begin{tabular}{|l|c|c|c|}
\hline
Method & Average F1 & Total Accuracy & Metric \\
\hline
Baseline\cite{2106.15318} & 0.30 & 0.50 & 0.366 \\
Ours & 0.402 & 0.630 & 0.477\\
\hline
\end{tabular}
\end{center}
\label{tab:expr}
\end{table}

\subsection{Results}

We use the cropped and aligned images provided in the Aff-wild2 dataset
when doing validation.
Some video frames are labeled, but there are no corresponding images
in the cropped and aligned folder. We discard these frames with their
labels when we evaluate our results on the validation set.

We achieve an AU score of 0.712 and an expression score of 0.554 on the validation set. See Table.\ref{tab:au} and
Table.\ref{tab:expr}.
The results show that our method greatly exceeds the baseline method.

\section{Conclusion and Future Work}

We proposed a multi-modal and multi-task learning method by using both visual and audio information for the competition of ABAW2021 in ICCV2021.
Our method obtained a score of 0.712 on AU recognition and 0.554 on expression recognition using the validation dataset. By using multi-modal information and multi-task
training method, the result of our approach far exceeding the baseline result.

For future work, we will analyze our approach and do more detailed
ablation studies to verify the principle that the method works.

{\small
\bibliographystyle{ieee_fullname}
\bibliography{egbib}
}

\end{document}